\documentclass[aps,prl,twocolumn,superscriptaddress,amsmath,amssymb]{revtex4-2}

\usepackage{graphicx}
\usepackage{booktabs}
\usepackage{array}
\usepackage{url}

\newcommand{\Complex}{\mathbb{C}}
\newcommand{\Repart}{\operatorname{Re}}

\newcommand{\concat}{\operatorname{concat}}

\begin{document}

\title{Phase-Associative Memory: Sequence Modeling in Complex Hilbert Space\footnote{Code and training logs available at \url{https://github.com/gowrav-vishwakarma/qllm2}.}}

\author{Gowrav Vishwakarma}
\email{gowrav@xavoc.com}
\affiliation{Xavoc Technocrats Pvt.\ Ltd., \url{https://xavoc.com}, India}

\author{Christopher J.\ Agostino}
\email{cjp.agostino@gmail.com}
\affiliation{NPC Worldwide, Bloomington, Indiana 47403, USA}

\date{\today}

\begin{abstract}
Experiments probing natural language processing by both humans and LLMs suggest that the meaning of a semantic expression is indeterminate prior to the act of interpretation rather than being specifiable simply as the sum of its parts (i.e. compositionality). This observer-dependent act dynamically actualizes meaning under genuine contextuality more consistent with quantum logical mechanisms than with classical Boolean approaches that assume separability, motivating an approach to language modeling that utilizes a Hilbert space formalism. In this work, we introduce Phase-Associative Memory (PAM)---a complex-valued sequence model whose state $S_t \in \Complex^{d \times d}$ accumulates outer products of complex token embeddings retrieved through the conjugate inner product $\mathrm{Re}\langle K \mid Q\rangle / \sqrt{d}$---and evaluate it against a structurally matched real-valued ablation. Both architectures train stably across a 5M--100M parameter sweep on WikiText-103 under identical conditions; PAM sits at higher absolute loss at every measured scale but improves more rapidly with parameter count, with power-law exponents of $-0.15$ vs.\ $-0.12$ in loss and $-0.65$ vs.\ $-0.49$ in perplexity that narrow the gap between the two architectures monotonically. Further investigation of complex-valued sequence modeling at larger scales could reveal that the loss plateau characteristic of real-valued state-of-the-art language models (e.g. transformers) is reachable with PAM-style architectures with an order of magnitude fewer parameters than the current frontier ($\sim$1T), implying that similar capabilities are achievable at sizes runnable on consumer-grade hardware.
\end{abstract}
\maketitle

\section{Introduction}
\label{sec:intro}

The assumption that a system's constituents can be analyzed independently of one another and of the conditions under which they are observed has been a foundational premise of empirical science since the seventeenth century~\cite{shapin1996scientific,dear2001revolutionizing}. Scholastic metaphysics, notably Aquinas~\cite{aquinas1274summa}, treated nature as composed of distinct substances knowable in isolation~\cite{crombie1959augustine}; early modern figures from Galileo~\cite{galileo1632dialogue} and Bacon (\textit{Novum Organum}~\cite{bacon1620novum}) through Descartes~\cite{descartes1641meditations} to Newton's \textit{Principia}~\cite{newton1687principia} sharpened a picture in which states and causes admit description without essential reference to the observer.

For more than two centuries classical physics reinforced separability and determinism as features of the world. Quantum mechanics challenged that picture, but its founders disagreed about what the resulting indeterminacy actually meant: Heisenberg argued that the act of measurement disturbs values existing independently of observation, Bohr contended that no such values exist prior to measurement and that measurement itself is what renders them definite, while Einstein, Podolsky, and Rosen maintained that separated systems must possess definite properties independent of measurement~\cite{einstein1935,howard1985einstein,howard1989holism}. Bell's theorem~\cite{bell1964,bell1966problem} was designed to test this disagreement, deriving inequalities that any theory of pre-existing, context-independent values must satisfy. The experiments that followed~\cite{clauser1969,aspect1982experimental} found these inequalities violated, with the observed correlations aligning more closely with Bohr's interpretation~\cite{bohr1935can} than with the realist alternative. The Kochen--Specker theorem~\cite{kochen1967problem} extended contextuality to single systems, and later tests closed remaining loopholes~\cite{hensen2015loophole,giustina2015significant,shalm2015strong,rauch2018cosmic}. Subsequent work in quantum information~\cite{deutsch1985quantum,shor1997polynomial,nielsen2000quantum} established inseparability as a resource, with the Tsirelson bound~\cite{tsirelson1980quantum} quantifying the quantum-over-classical advantage.

The study of language developed under the same assumptions of separability, though the connection is rarely made explicit. The principle of compositionality, which holds that the meaning of a complex expression is determined entirely by the meanings of its parts and the rules by which they combine~\cite{frege1892,montague1970}, treats semantic content as a property of linguistic constituents that can be analyzed independently of the interpreter and the context of interpretation. Whether this principle is adequate as a foundation for the study of meaning has been contested on philosophical grounds for more than a century~\cite{wittgenstein1953philosophical,quine1960word,gadamer1960truth}, but the computational study of language has largely proceeded as if it were settled. Zellig Harris showed that the distributional properties of words in a corpus could serve as a proxy for their semantic relationships~\cite{harris1954}, and this insight carried through the twentieth century into latent semantic analysis, word embeddings~\cite{mikolov2013,bengio2013representation}, and ultimately large language models~\cite{hochreiter1997long,devlin2019bert,brown2020language}. At each stage, the underlying computational assumption has been the same: words have meanings that can be represented as fixed points in a real-valued vector space, and the task of a model is to learn where those points are and how they compose. The transformer architecture~\cite{vaswani2017attention,bahdanau2015neural} is the most successful embodiment of this program.

Transformer-based large language models have largely succeeded in passing the well-established benchmarks of artificial intelligence and language understanding~\cite{brown2020language}. However, their adoption in domains that require guaranteed reliability has been hindered by persistent difficulties, most prominently hallucination~\cite{huang2023survey,ji2023survey} and susceptibility to prompt injection~\cite{perez2022ignore,greshake2023not}, which have resisted solution despite substantial engineering effort across architectures and scales. The improvements in capability that once accompanied increases in model size and training data~\cite{kaplan2020scaling,hoffmann2022training} have plateaued~\cite{muennighoff2025scaling,villalobos2025scaling}, and the frontier of the field has shifted toward test-time compute~\cite{snell2024scaling}, chain-of-thought reasoning, and agentic iteration as strategies for navigating the space of possible responses rather than producing the correct one directly. This shift is consistent with the observation that the informational burden of disambiguating a semantic expression grows superlinearly with its complexity~\cite{agostino2025quantum}, making the recovery of a single intended meaning from an expression of even moderate depth an intractable problem in the sense of relevance realization~\cite{vervaeke2013,jaeger2023}: the system cannot determine what is relevant from the input alone and must instead explore. Efforts to understand these systems through mechanistic interpretability, which attempts to decompose the internal representations of neural networks into individually meaningful components, have encountered difficulties that appear to be structural rather than merely technical. Sparse autoencoders trained on frontier models lose approximately 90\% of the model's capability when their reconstructions replace the original activations~\cite{gao2024scaling_sae}, extracted features have been shown to be ``neither selective nor independent'' when used for steering~\cite{mueller2024isolation}, and a recent review with twenty-nine co-authors described the field's foundational concepts as ``not yet established'' and its status as ``pre-paradigmatic''~\cite{sharkey2025open}. Theoretical work has demonstrated an exponential gap between the complexity of representing features in superposition and computing with them~\cite{adler2024complexity}, and sparse autoencoders have been proven to fail to recover ground-truth features except under conditions of extreme sparsity~\cite{cui2025limits}. The difficulty of producing reproducible, meaningful decompositions of neural network representations mirrors the experience of cognitive neuroscience, where decades of functional neuroimaging have shown that localized brain-behavior associations require sample sizes orders of magnitude larger than most studies have used~\cite{marek2022reproducible}, that seventy independent teams analyzing the same fMRI dataset reach substantially different conclusions about which brain regions are involved~\cite{botvinik2020variability}, and that inferring cognitive processes from regional activation is logically unreliable because brain regions participate in many functions simultaneously~\cite{poldrack2006can,button2013power}. In both cases, the assumption that the system can be understood by decomposing it into separable, localizable components has produced results that do not replicate.

In physics, the distinction between a decomposition that has not yet been found and one that cannot exist in principle, because the underlying properties are fundamentally indeterminate prior to measurement, is precisely what Bell's theorem was designed to settle, and the same framework can be applied to semantic interpretation. When the CHSH test is applied to human semantic judgments, the correlations between interpretations produced under different contextual framings violate the classical bound~\cite{busemeyer2012quantum,pothos2013can,aerts2009quantum,wang2014context,bruza2023contextuality,pothos2022quantum}, and when the same tests are applied to large language models trained on text that human cognition produced, the violations persist across four orders of magnitude in parameter count, with the distributional character of the contextuality orthogonal to every standard benchmark tested~\cite{agostino2025quantum,agostino2026production}. Sheaf-theoretic analysis of BERT's internal representations has identified over 77,000 instances of contextuality at the level of the embeddings themselves~\cite{lo2024contextuality,abramsky2011sheaf}. The non-separability is not confined to the behavioral outputs of these systems; it is present in the geometry of their learned representations~\cite{williams2025philosophy,chen2026entanglement}. Reconsidered under the premise that meaning is indeterminate prior to the act of interpretation and that natural language is semantically degenerate~\cite{agostino2025quantum}, it necessarily follows that hallucinations and jailbreaks are not anomalies to be eliminated but commonplace consequences of a system that interprets rather than retrieves~\cite{agostino2026production}. If the correlational structure of language is genuinely non-classical, the natural mathematical framework for describing it is the same one that was developed for quantum mechanics: a complex Hilbert space in which states carry phase, similarities are computed through the conjugate inner product, and interference between components is an intrinsic property of the algebra rather than a behavior that must be learned. Large language models built on real-valued representations and softmax attention may functionally replicate this structure, but they do so in the way that any classical simulation of a quantum system does: by using enough parameters to project the complex-valued correlations onto a real-valued space, at a cost in capacity and efficiency that grows with the complexity of the structure being represented.

The representation of signals in complex form has a long history in engineering and physics. \citet{gabor1946theory} introduced the analytic signal in his theory of communication, and \citet{oppenheim1981importance} demonstrated that phase carries more structural information than magnitude in both images and audio~\cite{oppenheim1981importance}. The geometric phase discovered independently by Pancharatnam in optics~\cite{pancharatnam1956generalized} and Berry in quantum mechanics~\cite{berry1984quantal} showed that phase relationships encode topological properties of the space traversed by a system, information that is lost entirely when the representation is projected onto real-valued magnitudes. Complex-valued neural networks have been developed along these lines for decades~\cite{hirose2012complex,arjovsky2016unitary,trabelsi2018deep,wisdom2016full,wolter2018complex}, and the holographic reduced representations introduced by Plate~\cite{plate1995holographic} demonstrated that complex multiplication and conjugation provide a natural algebra for binding and retrieving associations~\cite{gayler2003vector,kleyko2023survey1}. \citet{danihelka2016associative} incorporated this algebra into an LSTM with complex-valued cell states, and \citet{ramsauer2021hopfield} showed that the mathematical structure underlying softmax attention is a modern Hopfield network~\cite{hopfield1982neural,krotov2016dense} whose linear variant is the fast weight programmer~\cite{schmidhuber1992learning,schlag2021linear}. None of these efforts, however, has produced a complete language model that operates in complex space from embedding through retrieval to output at a scale where comparison with conventional architectures is meaningful.

Separately, the development of efficient alternatives to the transformer's attention mechanism has produced a body of work that provides the architectural scaffolding for such a model. The transformer~\cite{vaswani2017attention} computes attention as a softmax-normalized dot product between real-valued projections of the input, an operation that is powerful but quadratic in sequence length and requires a key-value cache that grows linearly during inference. Removing softmax yields a recurrence with matrix state $S_t = S_{t-1} + V_t K_t^\top$~\cite{katharopoulos2020transformers}, which \citet{schlag2021linear} showed is equivalent to the fast weight programmer introduced by Schmidhuber~\cite{schmidhuber1992learning}, an associative memory that accumulates associations via outer products and retrieves via matrix-vector product. Subsequent work has refined this structure in various ways: RetNet~\cite{sun2023retentive} adds exponential decay, GLA~\cite{yang2024gated} introduces data-dependent gating, DeltaNet~\cite{yang2024deltanet} replaces additive accumulation with a delta rule, and GateLoop~\cite{katsch2024gateloop} uses complex-valued gates. From the state-space model side, the Linear Recurrent Unit~\cite{orvieto2023resurrecting} established the importance of complex-valued diagonal recurrences for stable long-range modeling, Mamba~\cite{gu2023mamba} introduced input-dependent selection, Griffin~\cite{de2024griffin} validated gated linear recurrence at scale, and Mamba-2~\cite{dao2024transformers} proved the formal equivalence between structured SSMs and linear attention. From the LSTM lineage, mLSTM~\cite{beck2024xlstm} independently arrives at the same matrix-state recurrence, and RWKV~\cite{peng2023rwkv,peng2024eagle} has demonstrated this family at up to 14B parameters. With few exceptions, these models operate in real-valued space. \citet{ramsauer2021hopfield} showed that softmax attention implements a modern Hopfield network, and the linear variant of this associative memory is precisely the fast weight programmer that the matrix-state models generalize.

Operational quantum logic established that any system whose observables are contextual requires a non-Boolean algebraic structure naturally housed in a complex Hilbert space with the conjugate inner product~\cite{birkhoff1936logic,piron1964axiomatique,foulis1972empirical,coecke2001operational}. Bell-inequality tests applied to transformer-based language models yield violations~\cite{agostino2025quantum,agostino2026production}, indicating that real-valued architectures can approximate this non-classical correlational structure given sufficient parameters. Phase-Associative Memory (PAM) takes the matrix-state recurrence shared by the lineages described above and moves it into the space that operational quantum logic identifies as native to contextual systems. The state, keys, values, and queries are all complex-valued, and retrieval uses the conjugate inner product $K^* \cdot Q$ rather than the standard dot product, so that the selectivity of retrieval depends on the phase alignment between stored and queried representations.

The architecture emerged through a series of experiments in which each failure was informative. Early versions introduced tokens in complex phase space but destroyed phase information by passing representations through real-valued nonlinearities; correcting this with phase-preserving primitives materially improved results. A subsequent attempt to inject holographic key--value bindings into a vector-state SSM caused a regression in perplexity, because multiple bindings superposed in a single $d$-dimensional vector interfere destructively with the classical $O(1/\sqrt{n})$ capacity degradation~\cite{plate1995holographic}. PAM resolves this by upgrading the state from $\Complex^d$ to $\Complex^{d \times d}$, providing $O(d^2)$ associative capacity per head. The reported configuration interleaves channel mixing and PAM in each of 16 blocks with complex rotary position embeddings~\cite{su2024roformer} on queries and keys, and admits a dual computational form that is $O(T^2)$ for parallel training and $O(1)$ per token for recurrent inference with no KV cache.

The remainder of this paper proceeds as follows. The Method section specifies the PAM block, the complex primitives, and the training setup. The Results section presents the canonical 5M--100M parameter sweep against a structurally matched real-valued ablation (SAM), the phase structure of the learned complex embeddings, and the decoherence-gap argument that interprets the empirical real-valued loss floor as the diagonal projection of a complex von Neumann entropy. The Discussion examines the retrieval mechanism, the matrix state under training as decoherence in complex Hilbert space, and the loss-space crossover of the PAM and SAM scaling fits together with the computational cost of the architecture.

\section{Method}
\label{sec:method}

In this work, we evaluate whether a sequence model whose entire signal path operates in complex Hilbert space can scale competitively with real-valued architectures. To do so, we instantiate Phase-Associative Memory (PAM)---a complex-valued primitive that accumulates and retrieves token associations through the conjugate inner product---alongside a structurally matched real-valued ablation (SAM), and train both at five scales from 5M to 100M parameters under a single canonical configuration so that observed differences trace to architecture rather than tuning. The model consists of a complex-valued embedding layer, 16 identical blocks, and a tied complex output head. Each block applies channel mixing via a ComplexGatedUnit (CGU) followed by sequence mixing via a Phase-Associative Memory (PAM) layer, both with residual connections and learned scaling. All operations in the main signal path are complex-valued and phase-preserving; gates and decay parameters use real-valued projections over magnitude features, but the primary data path never converts complex representations to real-valued intermediate forms.

Complex quantities are represented as tensors with shape $[\ldots, d, 2]$, implementing $\Complex^{d}$ in split-real form. The complex linear map, given weight matrices $W_r, W_i \in \mathbb{R}^{m \times n}$, computes $y_r = W_r x_r - W_i x_i$ and $y_i = W_i x_r + W_r x_i$. The activation function is modReLU, $\operatorname{modReLU}(z) = \operatorname{ReLU}(|z| + b) \cdot z / |z|$ with learned bias $b$, which thresholds magnitude while leaving phase untouched. Normalization is RMS normalization applied to magnitudes with phase preserved: $\operatorname{ComplexNorm}(z) = s \cdot (|z| / \operatorname{RMS}(|z|)) \cdot z / |z|$ with learned scale $s$. The channel mixing layer (CGU) is a SwiGLU-style gating block in complex space:
\begin{equation}
  \operatorname{CGU}(z) = W_{\text{down}}\bigl( \mathrm{gate}_{\mathrm{phase}} \odot \operatorname{modReLU}(W_{\text{up}} z) \cdot \sigma(|W_g z|) \bigr)
\end{equation}
where the gate magnitude $\sigma(|W_g z|)$ controls how much signal passes and the gate phase controls what rotation is applied. Each of 16 blocks applies CGU then PAM with residual connections and learned scaling:
\begin{align}
  \tilde{z}^{(l)} &= z^{(l-1)} + \alpha^{(l)}_{\text{CGU}} \cdot \text{CGU}_l(\operatorname{ComplexNorm}(z^{(l-1)})), \\
  z^{(l)} &= \tilde{z}^{(l)} + \alpha^{(l)}_{\text{PAM}} \cdot \text{PAM}_l(\operatorname{ComplexNorm}(\tilde{z}^{(l)}))
\end{align}
where $\alpha^{(l)}_{\text{CGU}}$ is initialized to $1.0$ and $\alpha^{(l)}_{\text{PAM}}$ to $0.1$. Logits are computed via a tied complex inner product with the embedding table: $\text{logits} = z_{\text{out},r} \cdot E_r^\top + z_{\text{out},i} \cdot E_i^\top$.

\label{sec:pam}
PAM replaces both the recurrent backbone and the attention mechanism with a single module whose operations correspond directly to the quantum semantic framework described in~\cite{agostino2025quantum}. In that framework, a semantic expression $S_E$ is represented as a state vector $|\psi_{S_E}\rangle = \sum_i c_i |e_i\rangle$ in a complex Hilbert space, where the complex coefficients $c_i$ carry phase information with no classical analogue, and interpretation is the application of a Hermitian operator whose eigenstates represent possible meanings. PAM implements this structure computationally: tokens are embedded as complex vectors, associations between them are accumulated in a complex matrix state via outer products, and retrieval is the projection of a query onto the accumulated state through the conjugate inner product, the same operation that computes $P(m_i) = |\langle e_i | \psi_{S_E}\rangle|^2$ in the quantum semantic framework.

We use the term ``memory'' in this work in the sense established by the modern Hopfield network~\cite{ramsauer2021hopfield,hopfield1982neural,krotov2016dense}, the holographic reduced representations of~\citet{plate1995holographic}, and the fast weight programmer~\cite{schmidhuber1992learning,schlag2021linear}: the model's internal state functions as content-addressable associative storage of token-level bindings, accumulated by outer products and retrieved by inner-product similarity. \citet{ramsauer2021hopfield} showed that softmax attention itself implements a modern Hopfield network of this kind, so this framing places PAM within an existing lineage of attention-as-associative-memory rather than introducing a separate retrieval mechanism. The matrix state $S_t \in \Complex^{d \times d}$ is internal to the sequence-modeling primitive and is distinct from (i) retrieval-augmented generation~\cite{lewis2020retrieval} and other external document-store approaches that supplement a language model with a separate corpus, (ii) the transformer key--value cache~\cite{vaswani2017attention} that grows linearly with sequence length during inference, and (iii) the cognitive-psychology categories of episodic, semantic, or working memory, with which we make no claim of correspondence. The ``Phase-Associative'' prefix denotes that retrieval is the conjugate inner product $\mathrm{Re}\langle K^* | Q\rangle$ rather than the standard real dot product, generalizing associative recall from real to complex Hilbert space so that retrieval strength depends on the phase relationship between stored keys and queries.

Each PAM head $h$ maintains an independent complex matrix state $S_t^{(h)} \in \Complex^{d \times d}$, where $d$ is the head dimension. Unlike attention heads, which compute independent dot-product similarities over a shared representation, PAM heads are parallel associative memory banks: each accumulates its own key-value associations and retrieves independently, so the $H$ heads collectively maintain $H$ separate $d \times d$ memory matrices. The total state capacity across $H$ heads is $H \times d^2$ complex values per layer ($6 \times 64^2 = 24{,}576$ in our configuration). The input $x_t \in \Complex^D$ is projected into queries, keys, and values via a single complex linear map:
\begin{equation}
  [Q_t; K_t; V_t] = W_{\text{QKV}} x_t \quad \Rightarrow \quad Q_t, K_t, V_t \in \Complex^{H \times d} .
\end{equation}
Complex rotary position embeddings~\cite{su2024roformer} are applied to $Q$ and $K$ by multiplying each element by a precomputed unit-magnitude factor $e^{i m \theta}$, encoding absolute position in phase while leaving magnitudes unchanged; in the conjugate product $K_i^* \cdot \tilde{Q}_t$ the dependence on position difference $(m{-}n)$ yields relative position structure. Retrieval uses the scaled query $\tilde{Q}_t := Q_t / \sqrt{d}$.

The decay rate $\gamma_t$ controls how quickly the state forgets and is computed from the input as $\gamma_t = \exp(-\operatorname{softplus}(W_{dt} \cdot \concat(x_{t,r}, x_{t,i}) + b_{dt}))$, where $b_{dt}$ is initialized to $-4.0$ for slow initial decay. A learned protect gate $p_t = \sigma(W_p \cdot |x_t| + b_p)$ with $b_p = -3.0$ modifies the effective decay:
\begin{equation}
  \gamma_t = e^{-dt_t} \cdot (1 - p_t) + p_t, \qquad V'_t = V_t \cdot (1 - p_t) .
\end{equation}
When $p_t \to 1$ the state is frozen and new values are suppressed; when $p_t \to 0$ the decay proceeds normally. The state then evolves as:
\begin{equation}
  S_t = \gamma_t \cdot S_{t-1} + V'_t \otimes K_t^*
\end{equation}
where $\otimes$ denotes complex outer product and $K_t^*$ is the complex conjugate of the key. Retrieval computes $Y_t = S_t \, \tilde{Q}_t$, which expands to:
\begin{equation}
  Y_t = \sum_{i \leq t} \left(\prod_{j=i+1}^{t} \gamma_j\right) \bigl(K_i^* \cdot \tilde{Q}_t\bigr) \, V'_i .
\label{eq:pam-expand}
\end{equation}
The conjugate inner product $K_i^* \cdot \tilde{Q}_t$ determines retrieval strength through phase alignment: associations whose keys are phase-coherent with the query are retrieved strongly while phase-incoherent associations are suppressed, without softmax normalization. When the keys are mutually orthogonal, each outer product $V_i \otimes K_i^*$ occupies an independent subspace of the $d \times d$ matrix, so retrieval is lossless: $S \tilde{Q} = V_j$ exactly when $K_j^* \cdot \tilde{Q} = 1$ and $K_i^* \cdot \tilde{Q} = 0$ for $i \neq j$. This is in contrast to vector-state models, where superposed bindings in $\Complex^d$ interfere destructively with retrieval accuracy degrading as $O(1/\sqrt{N})$~\cite{plate1995holographic}. The matrix state supports up to $d$ lossless associations per head; the data-dependent decay then converts this from a lossless store into a controlled lossy one, where information loss is a learned gating decision rather than an unavoidable consequence of the storage format.

During training, the recurrence is computed in $O(T^2)$ time by forming a decay matrix $D \in \mathbb{R}^{T \times T}$ with $\log D[t,i] = \sum_{j=i+1}^{t} \log \gamma_j$ via cumulative sums, applying a causal mask, computing the complex score matrix $W = \tilde{Q} K^{*\top}$, and obtaining the output as $Y = (W \odot D) \cdot V'$. This is mathematically equivalent to the recurrence but parallelizes across the sequence dimension. During autoregressive generation, each token requires $O(Hd^2)$ work per layer, and the state $S \in \Complex^{H \times d \times d}$ is fixed-size and does not grow with sequence length.

We train and evaluate on WikiText-103~\cite{merity2017pointer}, approximately 103 million tokens of Wikipedia text tokenized with the GPT-2 BPE tokenizer (vocabulary size 50,257). The principal results of this work come from a five-scale sweep of the PAM and SAM architectures (5M to 100M parameters) under a single canonical training configuration; the per-scale architectural parameters are listed in Table~\ref{tab:config} and the shared training hyperparameters in Table~\ref{tab:train}. All blocks interleave a ComplexGatedUnit (CGU) with a PAM layer, with Gated State Protection enabled, complex RoPE applied to $Q$ and $K$, no QK phase normalization, and a CGU expansion factor of 3 throughout. SAM matches each scale's parameter count by using a wider real dimension and additional memory banks (configurations omitted for brevity; see release notes).

\begin{table}[t]
  \centering
  \caption{PAM scaling sweep architectures.}
  \label{tab:config}
  \begin{tabular}{@{}rcccc@{}}
    \toprule
    Scale & dim & layers & heads & head dim \\
    \midrule
    5M   &  44 & 12 & 2 & 16 \\
    10M  &  80 & 12 & 4 & 16 \\
    25M  & 200 &  6 & 4 & 16 \\
    50M  & 292 & 12 & 4 & 16 \\
    100M & 384 & 16 & 6 & 64 \\
    \bottomrule
  \end{tabular}
\end{table}

\begin{table}[t]
  \centering
  \caption{Canonical training hyperparameters (shared across all scales of the sweep).}
  \label{tab:train}
  \begin{tabular}{@{}ll@{}}
    \toprule
    Parameter & Value \\
    \midrule
    Optimizer & AdamW \\
    Learning rate & $3 \times 10^{-5}$ \\
    Weight decay & 0.01 \\
    Warmup steps & 500 \\
    LR schedule & warmup + cosine decay \\
    Batch size & 8 \\
    Sequence length & 512 \\
    Epochs & 10 \\
    Gradient clipping & 1.0 \\
    Hardware & Apple M4 Max \\
    Initialization & orthogonal (complex linear maps) \\
    \bottomrule
  \end{tabular}
\end{table}

Preliminary tests at different hyperparameters and on different hardware (NVIDIA RTX 4090 with bf16 mixed precision and \texttt{torch.compile}) showed qualitatively similar behavior to what we report from the canonical sweep. We restricted the principal computational runs to the configuration above for consistency across model sizes and architecture classes.

Generation samples are logged every 5,000 steps using temperature 1.0, top-$k$ 50, top-$p$ 0.9, and repetition penalty 1.2.

To examine the phase structure of the learned complex embeddings, we construct synonym, antonym, and random word-pair sets from WordNet~\cite{miller1995wordnet,fellbaum1998wordnet} restricted to lemmas that map to a single GPT-2 token. Synonym and antonym pairs are drawn from WordNet synsets and lemma antonym relations, with an equal-size set of random pairs sampled uniformly from the same single-token vocabulary. For each pair we compute the normalized conjugate inner product $\langle z_1^* | z_2 \rangle$ and report the joint distribution of its phase difference and coherence.

\section{Results}
\label{sec:results}

\label{sec:main-result}

In this section, we present the results of training PAM and its real-valued ablation SAM on WikiText-103. We first describe the training dynamics of the interleaved PAM configuration alongside several architectural ablations. We then fit power-law scaling relations for PAM and SAM across the 5M--100M sweep and compare them with published real-valued scaling laws. Finally, we characterize the learned complex embeddings of PAM through their phase structure on WordNet word-pair sets.

The interleaved PAM configuration trains stably on WikiText-103. An earlier sequential configuration (16 CGU layers followed by 16 PAM layers, no RoPE) underperforms it, and a hybrid that adds sparse windowed attention every fourth block produces no improvement, indicating that interleaving channel and sequence mixing matters and that supplemental attention provides no benefit at this scale. A variant with per-element unit normalization of $Q$ and $K$ before the conjugate inner product saw decreasing validation loss but collapsed into severe lexical repetition by mid-training and was stopped during epoch 5, indicating that both magnitude and phase must be free to vary for the retrieval mechanism to function.

\begin{figure}[t]
  \centering
  \includegraphics[width=\columnwidth]{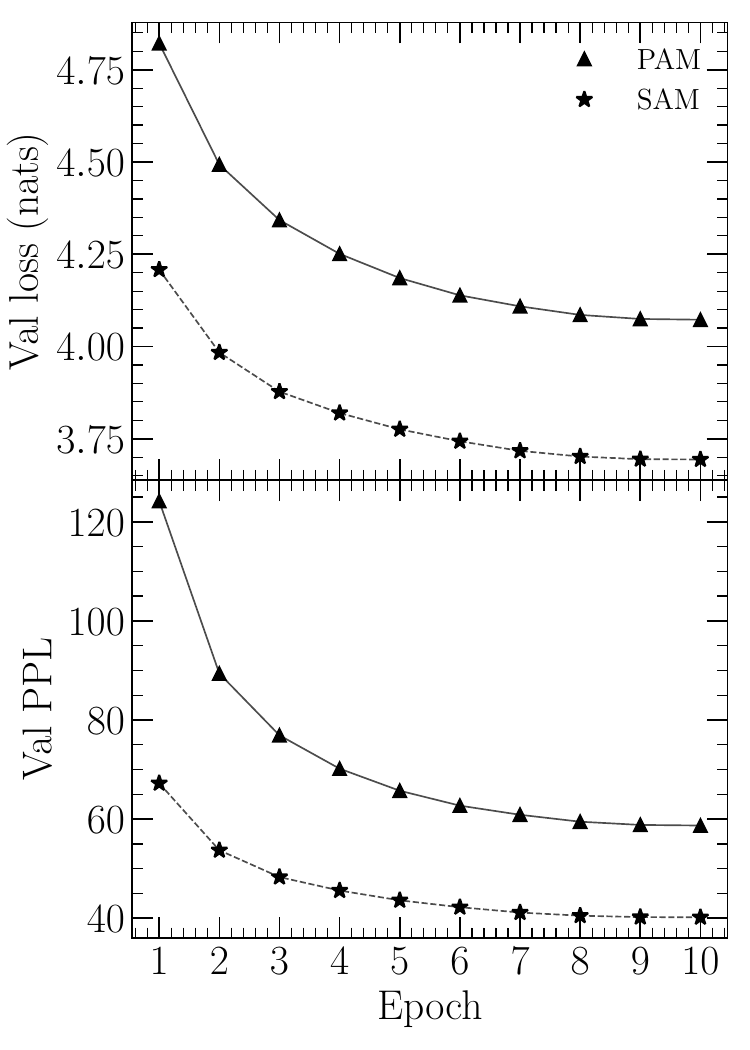}
  \caption{Per-epoch validation loss (top) and validation perplexity (bottom) at the 10M point of the PAM/SAM sweep on WikiText-103 (seq\_len 512, M4 Max). SAM converges faster and stabilizes lower at this scale (PPL 40.20 vs.\ 58.71); the scaling-law view in Figure~\ref{fig:scaling_loglog} shows that the gap narrows with model size.}
  \label{fig:cmp_val_ppl}
\end{figure}

At the 10M point of the sweep (PAM dim 80 with 4 memory banks, SAM dim 140 with 8), the per-epoch convergence trajectories (Figure~\ref{fig:cmp_val_ppl}) show SAM stabilizing earlier than PAM, consistent with the maturity of real-valued optimization.

The matrix state $S_t$ that accumulates these associations operates well below its theoretical capacity. The effective rank of $S_t$, measured via the entropy of its singular value distribution, saturates at approximately 10 out of $d = 64$ within the first 10--15 tokens and remains bounded thereafter. The learned decay keeps the state sparse, maintaining $\sim$10 active associations per memory bank at any given time. The $d^2$ lossless capacity of the matrix state sets the ceiling; the gated decay determines how much of it is occupied at each step.

To isolate the contribution of the complex formalism from the matrix-state architecture, we trained a real-valued variant (SAM) with architecturally identical structure but all complex operations replaced by real-valued equivalents. Because each complex linear map carries two weight matrices ($W_r, W_i$), SAM uses a wider dimension and additional memory banks at each scale to match the total parameter count of the corresponding PAM model. The only difference is the arithmetic: SAM uses the standard dot product $K_i \cdot Q_t$ for retrieval, real-valued outer products for accumulation, and ReLU-based activations in place of modReLU.

\begin{figure}[t]
  \centering
  \includegraphics[width=\columnwidth]{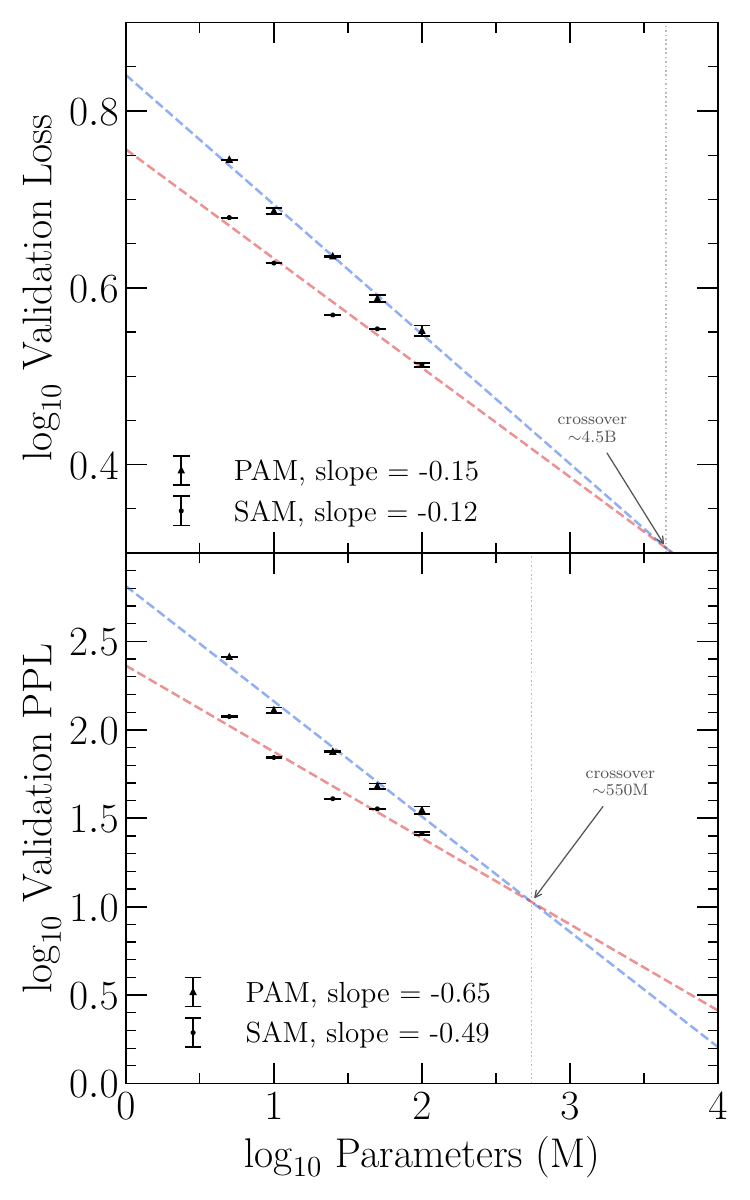}
  \caption{Scaling of validation loss (top) and validation perplexity (bottom) for PAM (triangles) and SAM (stars) on WikiText-103, in $\log_{10}$--$\log_{10}$ coordinates with shared $x$-axis. Each marker is the mean of late-epoch samples for one trained model; vertical bars are first-order propagated standard errors $\sigma_{\log_{10} y} = \sigma_y / (y \ln 10)$. Dashed lines are linear fits in log--log space, with slopes given in each panel's legend. Dotted vertical lines mark the extrapolated intersection of the two fits in each space: the loss-space fit crosses at $\sim$4.5B parameters (loss $\approx 2.05$ nats), the PPL-space fit at $\sim$550M parameters (PPL $\approx 10.6$).}
  \label{fig:scaling_loglog}
\end{figure}

We trained both PAM and SAM at five model sizes spanning 5M to 100M parameters under identical canonical conditions (lr $= 3 \times 10^{-5}$, batch size 8, sequence length 512, 10 epochs, M4 Max). For each trained model we report the mean validation metric over the late-epoch sample window (end-of-epoch and best mid-epoch evaluations from the last three epochs), with standard deviations propagated to log space via $\sigma_{\log_{10} y} = \sigma_y / (y \ln 10)$. Both validation loss and validation perplexity decrease monotonically with parameter count for both models (Figure~\ref{fig:scaling_loglog}). Linear regressions in $\log_{10}$--$\log_{10}$ coordinates --- fit on loss, the form directly comparable to standard neural scaling laws~\cite{kaplan2020scaling,hoffmann2022training}, and on perplexity --- give PAM a loss slope of $-0.15$ against SAM's $-0.12$, and corresponding PPL slopes of $-0.65$ against $-0.49$. SAM has lower absolute loss at every scale we measured (5.56 vs.\ 4.78 nats at 5M, 3.56 vs.\ 3.26 at 100M), but the gap narrows monotonically with scale, and the two fits intersect at $\sim$4.5B parameters in loss space (loss $\approx 2.05$ nats) and at $\sim$550M in PPL space (PPL $\approx 10.6$).

The 2.05-nat crossover lies close to the value the Kaplan power law predicts when extrapolated to 4.5B parameters on WebText2 ($\sim 2.12$ nats~\citep{kaplan2020scaling}), and is within $\sim 0.4$ nats of the Kaplan/Chinchilla irreducible-loss estimate of $\sim 1.69$ nats~\citep{hoffmann2022training}. GPT-2 1.5B reaches $\sim 2.3$ nats on WebText2~\citep{radford2019language}, Chinchilla 70B reaches $\sim 1.93$ nats~\citep{hoffmann2022training}, and GPT-3 175B reaches $\sim 3.0$ nats (PPL 20.5) on Penn Treebank~\citep{brown2020language}.

Assuming the formalisms of the quantum semantic framework~\citep{agostino2025quantum,agostino2026production}, in which semantic expressions are represented as states in a complex Hilbert space, the conditional state $\rho_{t \mid c}$ associated with a context $c$ governs the Born-rule probabilities for the next token $t$. A real-valued architecture with infinite capacity asymptotes at the Shannon entropy of $\rho_{t \mid c}$'s diagonal in the discrete-token basis, $H(\operatorname{diag} \rho_{t \mid c}) = -\sum_t \rho_{tt} \log \rho_{tt}$, since the only outputs it can produce are classical probability vectors over tokens. A Hilbert-space architecture can in principle represent the full conditional state and asymptotes at its von Neumann entropy, $S_{\text{VN}}(\rho_{t \mid c}) = -\operatorname{Tr}(\rho_{t \mid c} \log \rho_{t \mid c})$. The two are related by the elementary inequality $H(\operatorname{diag} \rho) \ge S_{\text{VN}}(\rho)$, with equality iff $\rho$ is already diagonal in the chosen basis, and the difference $\Delta_{\text{deco}} = H(\operatorname{diag} \rho) - S_{\text{VN}}(\rho) \ge 0$ is the relative entropy of decoherence, the information-theoretic cost of being structurally restricted to the diagonal of a state with off-diagonal coherences. We propose that one can interpret the empirical $1.69$-nat real-valued floor as the diagonal projection of the complex-valued von Neumann entropy of $\rho_{t \mid c}$ onto the classical subalgebra.

Given this framing, it necessarily follows that there should exist some gap in the irreducible loss between complex- and real-valued representations, with the size set by the structure of $\rho_{t \mid c}$. For a three-state conditional written as the rank-one projector onto a maximally-coherent superposition with non-trivial phases between basis vectors,
\begin{equation}
\rho_{t \mid c} \;=\; \frac{1}{3}\!
\begin{pmatrix}
1                  & e^{-i\pi/3}        & e^{-i 2\pi/3} \\[2pt]
e^{i\pi/3}         & 1                  & e^{-i\pi/3} \\[2pt]
e^{i 2\pi/3}       & e^{i\pi/3}         & 1
\end{pmatrix},
\end{equation}
the projector onto $|\psi\rangle = \tfrac{1}{\sqrt{3}}(|0\rangle + e^{i\pi/3}|1\rangle + e^{i 2\pi/3}|2\rangle)$. The diagonal is uniform, $\operatorname{diag}\rho_{t \mid c} = (\tfrac{1}{3},\tfrac{1}{3},\tfrac{1}{3})$, giving $H(\operatorname{diag}\rho) = \log 3 \approx 1.099$ nats --- the maximum Shannon entropy a real-valued architecture can assign to a three-outcome distribution. But $\rho_{t \mid c}$ is rank one, so $S_{\text{VN}}(\rho) = 0$: a Hilbert-space architecture that represents the full state assigns it zero entropy. The decoherence gap is the entire $\log 3$, and the whole prediction loss the real-valued architecture incurs for this state is information that lives in the off-diagonal phase coherences. Mixing this state with the maximally mixed one, $\rho(p) = (1-p)|\psi\rangle\langle\psi| + p\,\mathbb{1}/3$, gives eigenvalues $(1 - 2p/3,\, p/3,\, p/3)$, leaves the diagonal uniform, and interpolates the gap monotonically from $\log 3$ at $p = 0$ to $0$ at $p = 1$. More generally, the maximum decoherence gap for a $d$-outcome conditional is $\log d$, so anchoring at the empirical $E_{\text{real}}^{\text{Kaplan}} \approx 1.69$ nats per token gives
\begin{equation}
E_{\text{complex}}^{\text{floor}} \;\ge\; E_{\text{real}}^{\text{Kaplan}} - \log d_{\text{eff}},
\end{equation}
where $d_{\text{eff}}$ is the effective coherence dimension of the conditional state per token-prediction event: $d_{\text{eff}} = 2$ gives a predicted floor of $\sim 1.00$ nat per token, $d_{\text{eff}} = 4$ gives $\sim 0.30$ nat. Empirical estimates of $d_{\text{eff}}$ can be grounded in CHSH-style experiments on natural language interpretation~\cite{busemeyer2012quantum,pothos2013can,aerts2009quantum,wang2014context,bruza2023contextuality,pothos2022quantum,agostino2025quantum,agostino2026production}, which measure how many contextually-orthogonal meanings a semantic expression simultaneously supports under interpretation. Observed $|S|$ values in the range $2.0$--$2.6$ across both human and LLM experiments, together with the typical polysemy of content words, place $d_{\text{eff}}$ plausibly in the range 2--4 per token-prediction event, predicting a complex-valued floor between $0.30$ and $1.00$ nats per token---substantially below the $1.69$-nat real-valued figure, and indicating that the predicted gap is not merely formally positive but quantitatively significant. The strict positivity of the gap follows from QSF self-consistency alone, and the substantive prediction is that the $1.69$-nat figure in scaling-law analyses of real-valued transformers is not a fundamental floor on language modeling but the diagonal projection of a von Neumann entropy that a natively Hilbert-space architecture is in principle able to reach.

The architectural overhead that handicaps PAM at small scale --- the complex parameterization requires sufficient capacity to stabilize the conjugate-inner-product retrieval --- thus becomes a return on investment as scale grows, the pattern one would expect if the real-valued model is approximating, with classical machinery, a structure that the Hilbert-space model represents natively. With more careful optimization of the complex implementation (learning-rate schedules tuned for the modReLU activation, low-level kernel work) and training on corpora large enough that the upper end of the sweep stays within compute-optimal bounds --- the 50M and 100M points sit past Chinchilla-optimal for WikiText-103's $\sim$118M tokens even allowing for multi-epoch repetition --- we would expect the crossover transition to occur at smaller $N$ than this extrapolation predicts.

\begin{figure}[t]
  \centering
  \includegraphics[width=\columnwidth]{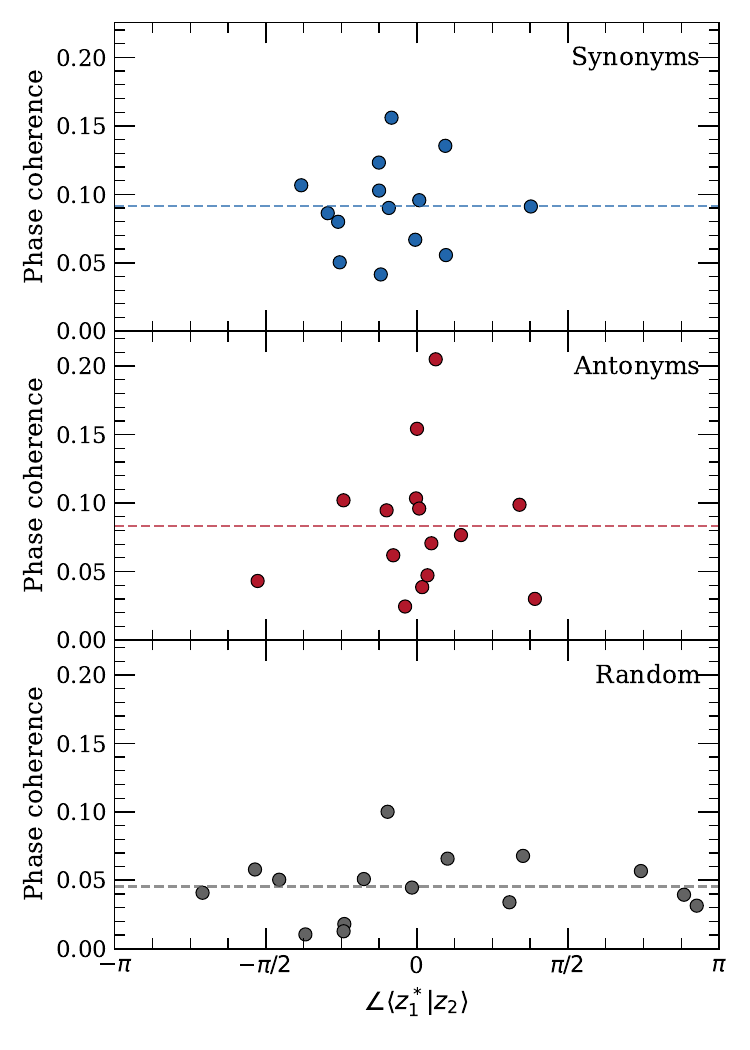}
  \caption{Phase coherence vs.\ phase difference $\angle\langle z_1^* | z_2 \rangle$ for synonym, antonym, and random word pairs in the learned complex embeddings. Dashed lines indicate group means.}
  \label{fig:phase_coherence}
\end{figure}

As a characterization of the learned complex embeddings, we examine their phase structure on a curated subset of WordNet pairs. Figure~\ref{fig:phase_coherence} shows that synonyms cluster near zero phase difference at elevated coherence in the conjugate inner product $\langle z_1^* | z_2 \rangle$, while unrelated pairs scatter across $[-\pi, \pi]$. No supervision over semantic relations was provided during training.

At epoch 10, given the prompt ``In 1923, the University of,'' the model generates: ``In 1923, the University of Illinois at Urbana @-@ Urdu said it was `an easy choice to do something in its own right.' The university also claimed the first students from Wisconsin had to be replaced by a more `good student' due to a lack of funds.'' The text shows some grammatical capability and avoids degenerate repetition (3-gram repetition rate 0.034, 4-gram repetition rate 0.011, unique token ratio 0.703), but is not factually reliable at this scale.

\section{Discussion}
\label{sec:analysis}

We examine three aspects of PAM's behavior in this section. We first analyze the retrieval mechanism, in which the conjugate inner product replaces the standard real dot product, and probe the role that magnitude and phase play in it. Next we explore the matrix state during training, where the effective rank stays well below capacity, and interpret this behavior through the lens of decoherence. Finally, we discuss the loss-space crossing of our PAM and SAM scaling fits, connect it to the irreducible-loss floor of language modeling, and assess the computational costs of the architecture.

\subsection{Retrieval through destructive interference}
\label{sec:dis_retrieval}

PAM addresses associative recall through destructive interference rather than nonlinear sharpening. The conjugate inner product $\Repart\langle K_i^* | Q_t\rangle$ takes negative or imaginary values when a stored key is phase-incoherent with the query, so the retrieval mechanism actively suppresses phase-mismatched associations rather than merely downweighting them. \citet{arora2024based} have shown that linear-attention models tend to struggle on associative recall because real-valued inner products are non-negative for matched directions, with every stored association contributing positively to a query and diluting the target signal; softmax attention escapes this through the exponential sharpening of its modern-Hopfield form~\cite{ramsauer2021hopfield}. Long-context passkey-retrieval experiments will settle directly whether destructive interference recovers specific associations from a large store as efficiently as exponential sharpening; we leave that test to future work.

If destructive interference is the retrieval mechanism, then PAM should depend on both the phase and the magnitude of $Q$ and $K$. We test this with the QK phase-normalization ablation. If the magnitude of $Q$ and $K$ were redundant given their phases, restricting the model to phase-only retrieval should leave training behavior largely unchanged. We find that this is not the case. The phase-only variant continued to drive validation loss down while generation collapsed into severe lexical repetition by mid-training, indicating that the directions in the loss landscape it was reaching were not generatively useful. We conclude that magnitude and phase carry distinct, non-redundant information in the retrieval process, and that removing the magnitude degree of freedom on its own collapses the generative signal entirely.

\subsection{Decoherence in the matrix state}
\label{sec:dis_decoherence}

The full matrix state $S_t \in \Complex^{d \times d}$ has $d^2$ complex degrees of freedom available for storage. If the model were filling that capacity straightforwardly, we would expect the effective rank of $S_t$ to grow with context length until it approached $d$ or some constant fraction of $d^2$. We find instead that the effective rank saturates at $\sim$10 out of $d = 64$ within the first 10--15 tokens of a sequence and remains bounded thereafter. The data-dependent decay drives most off-diagonal complex coherences to zero before they accumulate, and the rank stabilizes at the dimensionality of the small subset of associations relevant to the current context. In quantum-information terms~\cite{zurek2003decoherence} this is decoherence; in cognitive-science terms~\cite{vervaeke2013} it is relevance realization. We propose that these are two descriptions of the same operation, namely the selective suppression of structure that is not coupled to the present context.

\subsection{The loss-space crossover and computational costs}
\label{sec:dis_gap}

We apply the same decoherence framing to PAM's scaling behavior against the real-valued ablation. Our PAM and SAM fits cross in loss space at 4.5B parameters and $\approx 2.05$ nats, within $\sim$0.4 nats of the Kaplan/Chinchilla irreducible-loss estimate of $\sim$1.69 nats and within $\sim$0.07 nats of the Kaplan power law extrapolated to that scale on WebText2 ($\sim$2.12 nats). We acknowledge that 4.5B parameters lies far outside the regime that WikiText-103's $\sim$103M tokens can directly support, and that no model of that size is practically trainable on this corpus; we entertain the extrapolation as a proof-of-concept comparison against published real-valued scaling laws fit on much larger corpora, not as a literal forecast of where PAM and SAM would cross under fully resourced training.

Assuming the formalism of the quantum semantic framework, we interpret the empirical $\sim$1.69-nat real-valued floor as the diagonal projection of the complex von Neumann entropy of $\rho_{t\mid c}$ onto the classical subalgebra. CHSH violations of $|S| \sim 2.0$--$2.6$ on transformer-based language models force $\rho_{t\mid c}$ to retain off-diagonal coherences in any basis natural to language, so the relative entropy of decoherence between $H(\operatorname{diag}\rho_{t\mid c})$ and $S_{\text{VN}}(\rho_{t\mid c})$ is positive --- a gap that Hilbert-space architectures should reach below.

Beyond the asymptotic argument, on more practical grounds, computational costs are comparable to standard attention. The training-time dual form gives $O(T^2 H d)$ per layer, and at inference PAM uses a fixed state of 49,152 floats per layer regardless of sequence length, against a KV cache that grows linearly with context. At $T = 2048$ this state is $\sim$56$\times$ smaller than the transformer's KV cache, and the ratio grows with context length. Preliminary tests against matched dense transformers under the canonical configuration show comparably competitive results at the small scales we examined, with validation loss and perplexity values sitting between PAM and SAM. A thorough evaluation against the diversity of contemporary transformer implementations --- dense transformers across scales, mixture-of-experts variants, and alternative attention mechanisms --- will be the subject of further investigations that are beyond the scope of this work.

\section{Conclusion}

In this work, we have used a five-scale parameter sweep on WikiText-103 (5M to 100M parameters) to investigate the scaling behavior of Phase-Associative Memory (PAM), a language model whose representations and operations live in a complex Hilbert space, in comparison with a matched real-valued ablation (SAM). Our conclusions are the following:

\begin{enumerate}

\item PAM trains stably across the 5M--100M sweep on WikiText-103 (Figure~\ref{fig:scaling_loglog}) and reaches validation perplexity competitive with a structurally matched real-valued ablation under identical training, without optimization specialized to the complex arithmetic.

\item The matrix state $S_t$ accumulates associations well below its $d^2$ capacity. The effective rank, measured by the entropy of the singular-value spectrum, saturates at $\sim$10 out of $d = 64$ within the first 10--15 tokens and remains bounded thereafter. The gated decay determines occupancy.

\item PAM and SAM both show monotonic perplexity decrease with parameter count (Figure~\ref{fig:scaling_loglog}), but PAM's slope is steeper: $-0.15$ vs $-0.12$ in loss and $-0.65$ vs $-0.49$ in perplexity. The validation perplexity gap narrows monotonically from $2.18\times$ at 5M to $1.36\times$ at 100M.

\item Within the quantum semantic framework, we interpret the empirical $\sim$1.69-nat irreducible-loss floor characterized for real-valued transformer fits as the diagonal projection of the complex-valued von Neumann entropy of $\rho_{t \mid c}$. A natively Hilbert-space architecture can in principle reach below this floor, with the gap set by the structure of $\rho_{t \mid c}$. Anchoring the effective coherence dimension $d_{\text{eff}}$ in observed Bell-inequality violations of natural language interpretation places the predicted complex-valued floor between $0.30$ and $1.00$ nats per token.

\end{enumerate}

\bibliography{aps_refs}

\end{document}